\documentclass[final]{cvpr}

\usepackage{silence}
\usepackage{times}
\usepackage{epsfig}
\usepackage{graphicx}
\usepackage{amsmath}
\usepackage{mathtools}
\usepackage{amssymb}
\usepackage{multirow}
\usepackage{caption}
\usepackage{makecell} 
\usepackage{subcaption}
\usepackage{tabularx}
\usepackage[acronym]{glossaries}
\usepackage[utf8]{inputenc}
\usepackage{xr-hyper}
\graphicspath{{figs/},}

\usepackage[pagebackref=true,breaklinks=true,colorlinks,bookmarks=false]{hyperref}
\usepackage{cleveref}

\Crefname{equation}{Eq.}{Eqs.}
\Crefname{figure}{Fig.}{Figs.}

\newcolumntype{Y}{>{\centering\arraybackslash}X}
\newcolumntype{Z}{>{\raggedleft\arraybackslash}X}

\newcolumntype{L}{>{\raggedright\arraybackslash$}X<$}
\newcolumntype{M}{>{\centering\arraybackslash$}X<$}
\newcolumntype{N}{>{\raggedleft\arraybackslash$}X<$}

\newcolumntype{A}{>{\small}l}
\newcolumntype{B}{>{\small}c}
\newcolumntype{C}{>{\small}r}

\makeatletter
\newcommand*{\addFileDependency}[1]{
  \typeout{(#1)}
  \@addtofilelist{#1}
  \IfFileExists{#1}{}{\typeout{No file #1.}}
}
\makeatother

\newcommand{\multiline}[2]{\begin{tabular}{@{}#1@{}}#2\end{tabular}}

\newacronym{glidenet}{GlideNet}{Global, Local and Intrinsic based Dense Embedding Network}
\newacronym{fe}{FE}{Feature Extractor}
\newacronym{gfe}{GFE}{Global Feature Extractor}
\newacronym{lfe}{LFE}{Local Feature Extractor}
\newacronym{ife}{IFE}{Instance Feature Extractor}

\newacronym{car}{CAR}{Cityscapes Attributes Recognition}
\newacronym{vaw}{VAW}{Visual Attributes in the Wild}

\begin{document}

\title{
CAR -- Cityscapes Attributes Recognition\\ A Multi-category Attributes Dataset for Autonomous Vehicles
}

\author{Kareem Metwaly\thanks{Kareem Metwaly was in an internship at Scale AI through this work.}\\
Pennsylvania State University\\
Pennsylvania, USA\\
{\tt\small kareem@psu.edu}
\and
Aerin Kim\\
Scale AI\\
San Francisco, CA, USA\\
{\tt\small aerin.kim@scale.com}

\and
Elliot Branson\\
Scale AI\\
{\tt\small elliot.branson@scale.com}

\and 
Vishal Monga\\
Pennsylvania State University\\
Pennsylvania, USA\\
{\tt\small vmonga@engr.psu.edu}
}

\maketitle

\begin{abstract}
Self-driving vehicles are the future of transportation. With current advancements in this field, the world is getting closer to safe roads with almost zero probability of having accidents and eliminating human errors. However, there is still plenty of research and development necessary to reach a level of robustness. One important aspect is to understand a scene fully including all details. As some characteristics (attributes) of objects in a scene (drivers' behavior for instance) could be imperative for correct decision making. However, current algorithms suffer from low-quality datasets with such rich attributes. Therefore, in this paper, we present a new dataset for attributes recognition – Cityscapes Attributes Recognition (CAR). The new dataset extends the well-known dataset Cityscapes by adding an additional yet important annotation layer of attributes of objects in each image. Currently, we have annotated more than 32k instances of various categories (Vehicles, Pedestrians, etc.). The dataset has a structured and tailored taxonomy where each category has its own set of possible attributes. The tailored taxonomy focuses on attributes that is of most beneficent for developing better self-driving algorithms that depend on accurate computer vision and scene comprehension. We have also created an API for the dataset to ease the usage of CAR. The API can be accessed through \url{https://github.com/kareem-metwaly/CAR-API}
\end{abstract}


\section{Introduction}
\label{sec:introduction}

It is no secret that computer vision plays a vital role in many fields in life nowadays. One important field is autonomous vehicles. Autonomous vehicles are the future of transportation. First, they are much more reliable than humans, as a vehicle can make use of multiple sensors at once. For instance, a vehicle could use a Lidar \cite{li2020lidar} sensor to recognize far objects at night and mitigate an accident before it happens. Second, an autonomous car will have better and faster calculations of urgent situations. For example, a human suddenly falls in front of the vehicle.

There has been a vast amount of research work focusing on enhancing and reaching better and more reliable autonomous systems for driving. Primarily, existing work focus on object detection and classification. It is important to detect objects in a scene and classify them as well into different categories (pedestrian, vehicle, static object, movable object, etc). For instance, there has been a tremendous work on object detection using camera sensor \cite{ren2017faster, simonyan2015deep, DBLP:journals/corr/HasanPourRVS16}, Lidar \cite{qi2016pointnet, journals/corr/abs-1711-06396, lang2019pointpillars}, or both \cite{chen2017multiview, vora2020pointpainting, DBLP:journals/corr/abs-2004-12636}. However, complete understanding and comprehension of a scene requires more than detecting objects. We need to understand each object's attributes as well. The vehicles need to comprehend the surroundings with details to achieve reliable automation of transportation. For instance, an autonomous vehicle should distinguish between a green traffic light and a red one, between two types of traffic signs, and between an emergency vehicle and a normal one. Furthermore, Even if we recognize a speed limit traffic sign, we need to understand the value of the limit and identify the numbers. Attributes usually depend on the category of the object. A traffic light has a list of possible attributes that differentiates it from a pedestrian or a vehicle. Sometimes, attributes of an object depend on other objects in the scene, such as towing a vehicle. In contrast, others rely on the intrinsic properties of the object of interest.

Earlier methods for object detection and classification relied heavily on tailored or customized features that are either generated by ORB \cite{rublee2011orb}, SIFT \cite{lowe2004distinctive}, HOG \cite{dalal2005histograms} or other descriptors. Then, the extracted features pass through a statistical or learning module -- such as CRF\cite{lafferty2001conditional} -- to find the relation between the extracted features from the descriptor and the desired output. Recently, Convolutional Neural Networks (CNN) have proven their capability in extracting better features that ease the following step of classification and detection. This has been empirically proven in various fields, such as in object classification \cite{li2020group, huang2019convolutional}, object detection \cite{he2017mask, redmon2016yolo} and inverse image problems such as dehazing \cite{metwaly2020nonlocal, zhang2021learning}, denoising \cite{liu2021invertible,ren2021adaptive}, HDR estimation \cite{liu2020single, metwaly2020attention, chen2021hdrunet} ... etc. Having said that, learning techniques typically require a huge amount of data for training and/or regularization of the training such as \cite{cabon2020vkitti2, yu2020bdd100k, ancuti2020ntire, pougue2021debagreement, caesar2020nuscenes}. Utilizing non-learning based techniques to predict attributes of objects \cite{antwarg2012attribute, fang2010dangerous} is challenging and it generally produces lower performance compared to learning based techniques.

Moreover, scene comprehension requires distinguishing objects and understanding each object's different aspects (attributes). Two objects can belong to the same category, but their behavior may differ based on their intrinsic characteristics. To illustrate, suppose an autonomous vehicle driving on a highway, where there is another vehicle approaching fast and its driver is behaving irresponsibly. The autonomous vehicle has to notice the irresponsible behavior of the driver and perform some decisions (e.g. moving away from that lane or slowing down). An autonomous vehicle needs to recognize people with special needs as well, as those people may require more caution. For example, they may not hear the approaching vehicle or they may mistakenly and abruptly divert from the sidewalk to the street. This holistic and detailed information is crucial for life-threatening situations. Thus, we not only need to detect and classify objects in a scene into categories but we also need to classify their attributes as well.

Commonly, we define attributes as semantic descriptions of objects in a scene. Semantic information of an object incorporates how it looks, how it interacts and behaves, and how other objects affect and influence it. In general, the set of possible attributes of an object depends on the category of that object. For instance, a table might have attributes related to shape, color, and material. However, a human will have a completely different set of attributes related to age, gender, and activity status (sitting, standing, walking, ... etc). Though, some attributes may exist between multiple categories such as the visible percentage of an object. For consistency through the paper, we differentiate between classes of objects or classes of attributes by using two separate terms. A category refers to a certain class of objects; for example a vehicle category, a pedestrian category or a static object category. On the other hand, we use an attribute class or simply an attribute to refer to a class of attributes; for example a visibility attribute, a walking attribute and so on.

To predict attributes of objects correctly, we need to consider the following:  1) some attributes are specific to certain categories, 2) some categories may share the same attribute, 3) some attributes require a global understanding of the whole scene and 4) some attributes are intrinsic to the object of interest. To the best of our knowledge, most existing datasets do not have a structured taxonomies of attributes where the set of attributes of an object depends on its category. Moreover, existing datasets for autonomous vehicles do not focus on attributes and they either focus on categories or bounding boxes. In this paper, we present a new dataset \gls{car}, which is a multi-category attributes dataset for autonomous vehicles. \gls{car} focus on a new aspect yet vital for autonomous vehicles.


\section{Related Work}
\label{sec:related_work}

Attributes prediction shares some similarities with other popular topics in research such as object detection \cite{wang2021end, joseph2021towards}, image segmentation \cite{huynh2021progressive, li2021semantic} and classification \cite{liu2021ntie, srinivas2021bottleneck}. However, visual attributes recognition has its unique characteristics and challenges that distinguish it from other vision problems such as multi-class classification \cite{reese2020lbcnn} or multi-label classification \cite{durand2019learning, chen2019multi}. Examples of these challenges are as follows. First, the number of attributes to predict is usually large, unlike objects categorization, as some attributes are detail-oriented. Second, attributes usually depend on the category of the object. For example, a static object, such as a pole, can't walk or move. Lastly, incorporating global context is sometimes required to understand certain attributes, as some attributes depend on the interaction between objects. For example, to recognize a car towing another, we need to understand the existence of two cars and that they are connected; focusing on the towed car by itself would not let us understand that it is being towed. That has motivated recent several previous research to investigate how to tailor a recognition algorithm that can predict attributes such as \cite{pham2021learning, sarafianos2018deep, metwaly2022glidenet, jiang2020defense}. However, most related work focuses on either a small set of generic attributes \cite{kalayeh2021symbiosis, wang2017joint, tay2019aanet, rothe2015dex, li2016human, wang2021pedestrian} or a set of attributes of a specific category \cite{he2017adaptively, park2018attribute, yang2020hierarchical, tang2019improving, li2018landmark, abdulnabi2015multi}. For instance, \cite{huo2016vehicle, sun2019vehicle} attempt to predict attributes of vehicles. Sun \etal \cite{sun2019vehicle} proposes an algorithm for the prediction of vehicle attributes. Their proposed method is very specific to predicting the brand and the color of the vehicle. They use two parallel neural networks, one for each attribute. A combined learning schedule is utilized to train the model on both attributes. On the other hand, Huo \etal \cite{huo2016vehicle} use a convolution block first to extract important features, then a multi-task stage is used that is primarily a fully connected layer per attribute. The output of each fully connected layer is represents the probability of a particular attribute. On the other hand, \cite{abdulnabi2015multi, jia2020rethinking, tang2019improving} tackle the prediction of attributes related to pedestrians or humans. Jahandideh \etal \cite{jahandideh2018physical} attempts to predict physical attributes such as age and weight. They use a residual neural network and train it on two datasets; CelebA \cite{liu2015faceattributes} and a self-developed one \cite{liu2015faceattributes}. Abdulnabi \etal \cite{abdulnabi2015multi} learns semantic (visual) attributes through a multi-task CNN model, each CNN generates attribute-specific feature representations and shares knowledge through multi-tasking. They use a group of CNN networks that extract features and concatenate them to form a matrix that is later decomposed into a shared features matrix and attribute-specific features matrix. Although the idea is plausible but it requires a significant amount of resources and it is challenging to perform real-time prediction. Zhang \etal \cite{zhang2020solving} attempt to focus on datasets with missing labels and attempt to solve it with ``background replication loss''. Multiple datasets focus on attributes of humans, but the majority target facial attributes such as eye color, whether the human is wearing glasses or not, etc. Examples of datasets for humans with attributes are CelebA \cite{liu2015faceattributes} and IMDB-WIKI \cite{rothe2015dex}. Li \etal \cite{li2019visual} propose a framework that contains a spatial graph and a directed semantic graph. Using the Graph Convolutional Network (GCN), we could capture the spatial relationships between the regions and the semantic relationships between attributes as well.

Despite its importance, few existing works deal with a large and diverse amount of attributes and its categories \cite{pham2021learning, huang2020image, metwaly2022glidenet, yang2020hierarchical}. Sarafianos \etal \cite{sarafianos2018deep} proposed a new method that targeted the issue of class imbalance. Although they focus on human attributes, their method can be extended to other categories as well. On the other hand, Pham \etal \cite{pham2021learning} proposed a more generic multi-category attribute prediction method. It utilizes a spatial attention scheme that combines multi-level features to have a final feature vector that is later used for attribute prediction similar to how multi-label classification works. To achieve this, they utilized the GloVe \cite{pennington2014glove} word embedding. However, their method cannot be directly applied to complex taxonomies where the desired attribute vectors depend on the categories of the objects. Progressively, \cite{metwaly2022glidenet} proposes another multi-category attributes prediction where they use a new convolutional layer dubbed informed convolution to pay attention to the object of interest during the extraction of features. Their method can handle a variable number of attributes, making it more realistic in applications and yielding impressive results in the \gls{car} dataset. Ni and Huttunen \cite{ni2021vehicle} have a very good survey of recent work in attributes prediction of vehicles.

We believe the main problem with attributes prediction, or any other vision problem per say, is the availability of high-quality datasets to utilize. That has led many research to focus on producing good quality datasets such as NEURIPS Data-Centric AI workshop \cite{neurips2021datacentricai}. Some existing datasets for vehicle attributes recognition (e.g. color, type, make, license plate, and model) can be found in \cite{yang2015large, liu2016deep} but they usually lack a complex taxonomy of a wide range of categories with a vast amount of attributes. Not only that, they usually do not have panoptic segmentation \cite{kirillov2019panoptic} of different objects in each image, which is of foremost importance for a complete scene understanding. Pham \etal \cite{pham2021learning} proposed a new dataset \gls{vaw} that is rich with various types of categories where each object in an image has three sets of attributes; positive, negative, and unlabeled attributes. Their dataset is based on VGPhraseCut \cite{wu2020phrasecut} and GQA \cite{hudson2019gqa}. However, because VAW dataset is general and does not focus on a single domain, it falls short in a few areas for a specific use case, such as autonomous driving.  Colors and texture make up the majority of VAW's attributes, which are less critical in autonomous driving. The attributes that matter in autonomous driving are, for example, visibility (occlusion), parking status, the indicator light on/off, etc.


\section{Criteria for Autonomous Driving Dataset}
\label{sec:criteria_dataset}

In this section, we discuss the criteria we considered before the labeling process. First, we could either start from scratch by our own images, or base our attributes labeling on previous open-source available dataset. We chose the latter for two reason. First, there is already a huge amount of available images that we can utilize. Second, there is a vast amount of different datasets that by adding attribute labels to it, it gets a step closer to having complete scene information.

After we decided to chose a pre-existing images, we had several options for the dataset to label. We are only interested in datasets that have rich types of annotations since adding attributes to them would further make them complete and also encourages the development of new algorithms that can utilize additional information from different annotations. The following criteria is developed to appropriately choose the most suitable dataset.
\begin{itemize}
    \item The dataset must have at least $10$ classes.
    \item It must at least contain these classes: pedestrian - vehicles - traffic lights - traffic signs.
    \item It must contain Panoptic Segmentation (binary masks of instances for the aforementioned classes). This enforces the importance of the developed dataset as it may now contain attributes related to traffic signs and traffic lights, which does not exist till now to the best of our knowledge.
    \item It must contain 3D box annotations. We are interested in complementing existing datasets. If we choose a dataset with a limited type of labels, it would not be as useful in complete scene understanding.
    \item It must have at least $10^6$ instances. The number of instances should be large to consider all possible cases for all attributes over all categories. In addition, all images should be completely labeled; there shouldn't be missing labels or neglected instances.
    
    \item It must have real images not synthetic images generated by computers. At the end, this dataset would be used in real-life situations. Working with such synthetic images may lead to unexpected behaviors with some algorithms that may not make it reliable for benchmarking.
\end{itemize}

Taking the aforementioned criteria into considerations, we did not choose any of the following datasets.
\begin{itemize}
    \item BDD100k \cite{yu2020bdd100k}, CamVid \cite{brostow2008segmentation}, $D^2$-City \cite{che2019d2}, KAIST \cite{choi2018kaist}, IDD \cite{varma2019idd} and Mapillary Vistas \cite{neuhold2017mapillary}: while these datasets have a reasonable good amount of images they do not have 3D Boxes. Even though adding attributes to them may enrich it, but having a dataset with a richer set of labels is more favorable.
    \item Waymo Open \cite{sun2020waymo}, A*3D \cite{pham2020astar3d}, Lyft \cite{kesten2019level5}, Boxy \cite{behrendt2019boxy} and Agroverse \cite{chang2019agroverse}: these datasets do not have instance masks. This is vital for developing sophisticated algorithms which may want to use the binary mask of instances to pay more attention for better attributes prediction. For instance, \cite{metwaly2022glidenet,pham2021learning} use the binary mask in attributes prediction.
    \item Synscapes \cite{wrenninge2018synscapes}, Synthia \cite{ros2016synthia}, V-KITTI \cite{gaidon2016virtual}, V-KITTI 2 \cite{cabon2020vkitti2}, VIPER \cite{richter2017playing} are synthetic datasets. These datasets would not be as reliable as real datasets. In addition, they will increase the burden over developing algorithms that may apply what is learned through synthetic data over real situations.
    \item KITTI \cite{geiger2012kitti}, H3D \cite{patil2019h3d} and AS LiDAR \cite{ma2019trafficpredict} have small number of categories which makes it very limited and not useful for complete scene understanding. Specifically, KITTI and H3D have 8 categories, while AS LiDAR has 6 instead.
    \item COCO \cite{lin2014microsoft} is a great dataset that has many types of annotations. However, we had to not choose it as not all images are fully annotated and it would take effort to clean out images with missing labels. In addition, it does not contain 3D box annotations as well.
\end{itemize}
 
\begin{table*}
    \caption{Comparison between possible datasets in terms of size, available types of labels, and number of classes}
    \label{tab:datasets_comparison}
    \begin{tabularx}{\linewidth}{l||Y|Y|Y|Y}\Xhline{4\arrayrulewidth}
         & Cityscapes \cite{cordts2016cityscapes} & NuScenes \cite{caesar2020nuscenes} & Apolloscape \cite{wang2019apolloscape} & A2D2 \cite{geyer2020a2d2} \\\Xhline{2\arrayrulewidth}
        Cited by & $\sim 5318$ & $\sim 774$ & $\sim 210$ & $\sim 53$ \\\hline
        Year & $2016$ & $2019$ & $2020$ & $2020$ \\\hline
        \# images & $5k(+20k)$* & $40k^+$ & $70k^\ddagger$ & $12k^\dagger$ \\\hline
        \multiline{l}{Available\\Annotations} & \multiline{c}{3D Boxes\\Images Panoptic Seg.} & \multiline{c}{3D Boxes\\LiDAR Panoptic Seg.$^+$} & \multiline{c}{3D Boxes\\Lane Segmentation\\Car/Person Seg.} & \multiline{c}{3D Boxes\\Images Panoptic Seg.$^+$}\\\hline
        \# Categories & 23** & 10** & ~20 & 30 \\\hline
        \# Attributes & 5** & 4** & ~4 & 3 \\\hline
        \multiline{l}{Location\\of images} & 50 cities & \multiline{c}{Boston\\Singapore} & 4 China & 3 Germany \\\Xhline{4\arrayrulewidth} 
    \end{tabularx}
    \small{\textit{
    $*$ there are 5k images with fine-annotations and additional 20k images with coarse-annotations.\\
    $**$ the official numbers are different (23 category and 12 attributes for NuScenes and 32 category with no attributes for Cityscapes). However, the taxonomy and definition of each category differ from a dataset to another. Thus, we present numbers based on combining some categories into a single one while extending the attributes set.\\
    $+$ Although they have 1.4M images, but actual annotated frames are 40k. NuImages (a more recent version) have 93k images \cite{fong2021panoptic}.\\
    $\dagger$ They have about 40k images with Panoptic Seg. but only 12k include 3D boxes.\\
    $\ddagger$ They have more annotated frames (140k in total) without 3D boxes.
    }}
\end{table*}

After eliminating the previously mentioned datasets for there respective reasons, we ended up with other datasets shown in \Cref{tab:datasets_comparison} to choose from. In \Cref{tab:datasets_comparison}, we present a comparison between different datasets from different aspects. First, the citation number indicates the importance of a dataset in the research community. A well-cited dataset indicates that it is vastly used and having extra label layer over it would help in developing already existing algorithms as well as building new ones. Second, the number of images in the dataset is certainly important. Larger number of images would help in better learning algorithms. Third, all datasets have 3D box annotations. In Cityscapes, the 3D box annotation was added later by G\"ahlert \etal in \cite{gahlert2020cityscapes}. Apolloscope contains lane segmentation which is a good addition towards complete scene understanding. However, it does not have panoptic segmentation of different objects in each image. It only contains segmentation of vehicles and pedestrians. 

In the comparison, we have adjusted the number of categories and attributes to make it consistent over all datasets. As the definition of a category or an attribute varies from one dataset to another. For instance, we consider only two categories of vehicles; namely medium-to-large vehicles and small vehicles. Then in the former one, we have an attribute regarding the form, e.g. sedan, hatchback, SUV, van, truck ... etc. While in the latter, we have an attribute for the type, e.g. bicycle, motorcycle, flat drivable Surface ... etc. More discussion about the datasets and the selection criteria is found in \Cref{sec:supp_discussion_datasets}.

At the end, we decided to focus on Cityscapes dataset. First, it has a considerably large number of categories. Second, it has been well cited in research. Adding attributes to it will help in completing the scene understanding and developing better algorithms and easily compare it with previously developed ones. Third, it contains panoptic segmentation and 3D box labeling which is not found in most of the available datasets. Finally, it contains images from $50$ different cities. This makes its images very diverse in nature and have plenty of different situations and different cases for the attributes that we will be interested in labeling.


\section{\acrfull{car}}
\label{sec:car}

\gls{car} splits object's into two different groups; namely ``Things'' and ``Stuff'' similar to \cite{huang2021crossview,kirillov2019panoptic}. The type ``Things'' contains all countable types of objects, e.g. vehicles, pedestrians. On the other hand, the type ``Stuff'' contains other uncountable objects, e.g. sky, road, vegetation. Usually with Stuff objects, we are not concerned with their counts (even if this is possible), we are only concerned with some attributes related to it. For instance, we are interested in differentiating between a terrain vegetation or trees. As an exemplary scenario, imagine an autonomous vehicle that needs to move away of the road suddenly (due to a sudden object for instance in front of it that it cannot evade). It needs to distinguish between a terrain that it can drive over or a tree that will be considered another obstacle.

To further boost the usage of the dataset and enable simple usage of the dataset with other frameworks such as PyTorch \cite{paszke2019pytorch} and Tensorflow \cite{tensorflow2015}, we have developed an API \footnote{CAR-API can be accessed through this link: \url{https://github.com/kareem-metwaly/car-api}}\cite{car_api} that can be used to download and use the data with PyTorch. The API can be easily extended to other frameworks, but it may need some coding.

An example in the taxonomy of the dataset is that for objects of category ``Pedestrian'' which has $10$ attributes and belongs to ``Things'' type of categories, thus they are segmented per instance level. It contains a ``Visibility'' attribute to distinguish how much of the pedestrian is visible. This is important in some applications such as object tracking. We divide it into 5 classes; divided uniformly from $0\%$ to $100\%$. The second attribute relates to the ``Age''. It can only take one of two values; either Adult or Child. In some cases, it may be important to recognize the existence of children in the vicinity of the vehicle; to drive slower or be extra cautious of their unexpected behaviors. Another attribute is the ``Pedestrian Type'' which indicates whether the human is a ``Police Officer'', ``Construction Worker'' or ``Neither''. Sometimes, the vehicle may need to switch from lane where a police officer or a construction worker is standing close to it. The ``Activity'' attribute indicates whether the human is ``Sitting'', ``Lying Down'', ``Standing'', ``Walking'', ``Running'' or ``Riding''. For other attributes of this category and others, the reader is encouraged to check \Cref{sec:supp_taxonomy} of the supplementary document where it discusses the complete taxonomy of \gls{car} dataset.

\begin{figure}
    \centering
    \includegraphics[width=0.8\linewidth]{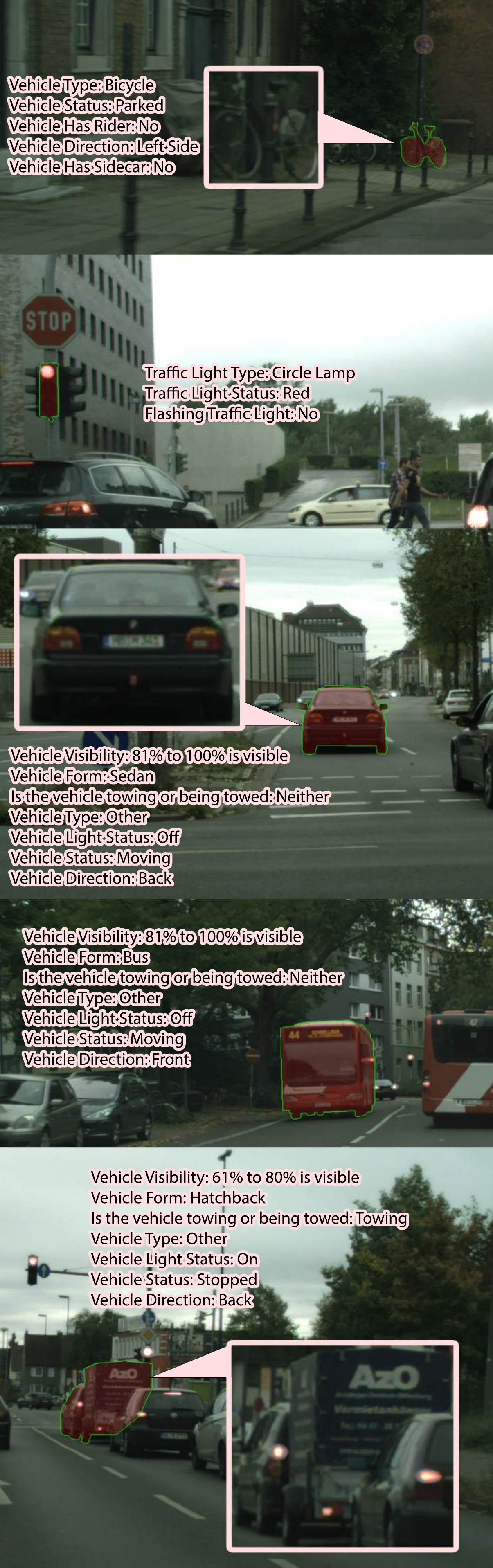}
    \caption{Examples of annotated instances from different categories and their respective attributes values}
    \label{fig:examples_annotations}
\end{figure}

\Cref{fig:examples_annotations} shows annotation examples from \gls{car} Dataset. As it can be noticed, the set of attributes change by changing the object category. The tailored taxonomy serves developing algorithms that better comprehend the scene and its objects. Some attributes are challenging to estimate considering the object of interest solely, such as whether a vehicle is parked or moving. For this reason we show the whole image to annotators so that they can better estimate the attributes of the object while at the same time highlighting the object. We always keep an ``Unclear'' value for annotators as an option to make sure that the results are as accurate as possible rather than giving a false-value that may damage developing algorithms and techniques.

\subsection{Labeling Procedure}\label{subsec:label_procedure}
\begin{figure*}
    \centering
    \includegraphics[width=\textwidth]{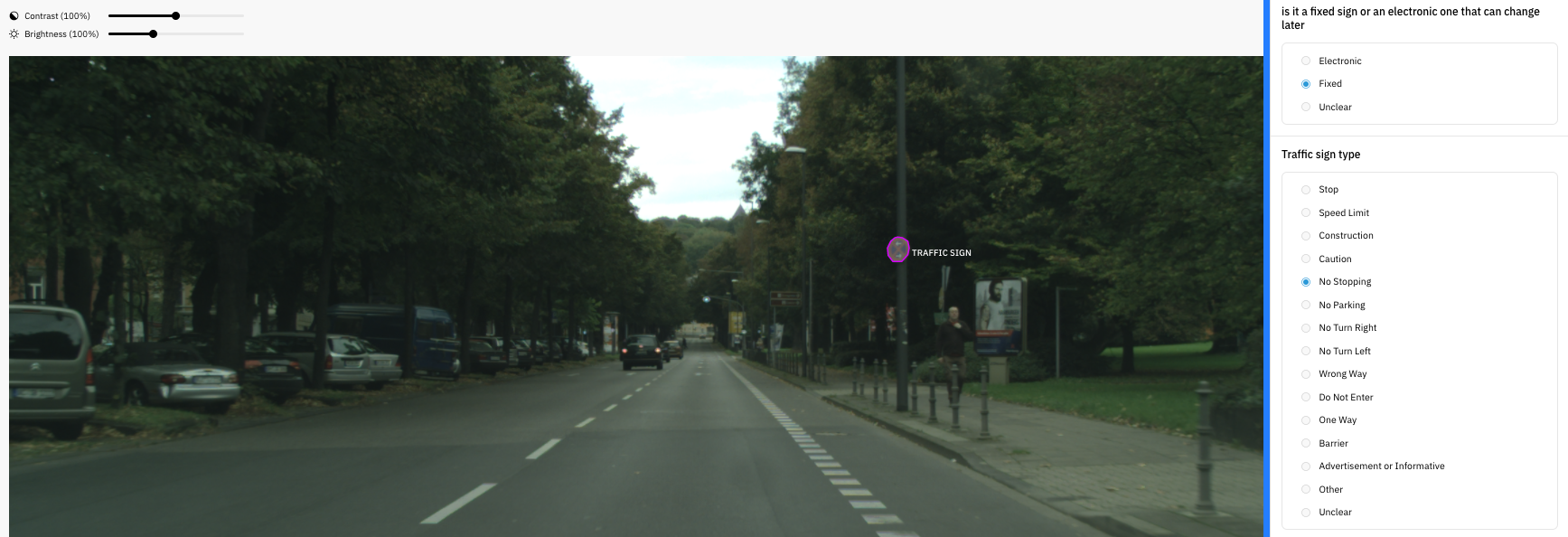}
    \caption{Example of the user interface for annotators}
    \label{fig:label_example}
\end{figure*}
The desired taxonomy of labels is complicated. Unlike most datasets, we have attributes that depend on the category and attributes that depend even on the value of other attributes. For instance, the category ``Traffic Sign'' has ``Type'' and ``Value'' attributes. The type of a traffic sign simply defines whether it is a stop sign, a speed limit sign, etc. The value however is a conditional attribute where its value depends on the type. For example, the value of a speed limit traffic sign will be the speed limit. The value of a detour traffic sign will be a pointer forward, right, etc. Therefore, it is not easy to perform labeling accurately and swiftly without designing a pipeline to assure the quality of the labeling. In this subsection, we discuss the details of the labeling procedure and how we make sure a minimum level of quality is guaranteed.

We believe that having low number of labels with accurate labels is significantly better than a larger number with inaccurate labels. This removes the burden of developing algorithms that can deal with inaccurate labels. In addition, this ensure that a test sample does actually represent a real-case situation. Therefore, rather than counting on a single person (annotator) to label each instance, we perform a consensus of 5 annotators. Not only that, but we also guarantee the quality of each annotator by assigning a quality metric to them. The quality metric is calculated by giving the annotator instances with labels that are known to us but hidden to them. These instances are fed randomly to the pipeline of the annotator to make sure they do not predict whether an instance is a test task or not. Finally, we check whether their labels matches the correct known labels and based on the response we assign the aforementioned metric. It is important to notice that even though we are using a consensus of 5 annotators, we do not combine their contributions equally. We combine it based on the quality metric. The framework uses Scale AI API for sending new tasks for annotation.

We also use the quality metric to ensure that annotators fully-comprehend the desired output labels as some annotators will start labeling without paying attention to the guidelines document. In addition, the quality metric is used to eliminate any adversarial annotators who may negatively impact the labeling quality by setting threshold over the quality metric.

\Cref{fig:label_example} depicts an example of what the annotator sees during labeling procedure through Scale AI platform. Depending on the category type of the object, a set of questions appear and a polygon highlighting the object of interest in the image as well as its category name. The annotator can play with the brightness and contrast to be able to better understand the attributes of the object. Further, the annotator can control the visibility of the polygon to make sure that no important information is obscured.

\subsection{\gls{car} Analysis}

\Cref{tab:attributes_per_category} shows some statistics of each category of the labeled instances. The reader is encouraged to check \Cref{sec:supp_taxonomy} in the supplementary document for complete information about the taxonomy; which entails the attributes structure, their category-dependency and the possible values each attribute can take. Unlike previous attributes datasets (such as \gls{vaw}), we do not consider binary value attributes. However, each attribute in \gls{car} can have one of many possible values. For example, ``Vehicle Type'' attribute can take one value of the following: Police Vehicle, Ambulance, Fire Vehicle, Construction Vehicle, Neither, or even Unclear. This makes the possible number of combinations of attributes an object can have to be very large. Therefore, we show in \Cref{tab:attributes_per_category} two values that can be used to understand the richness and diversity of attribute labels of each object category. First, $\mu_a$ represents how many possible values an attribute of each category can have \Cref{eq:avg_attr_vals}. On the other hand, $\pi_a$ represents how many unique combinations of values of attributes we can have \Cref{eq:attr_comb}. Mathematically they are expressed as follows.
\begin{tabularx}{\linewidth}{YY}
\begin{equation}
\mu = \frac{1}{n_c}\sum_{a=1}^{n_c} v_a \label{eq:avg_attr_vals} 
\end{equation} &
\begin{equation}
\pi = \prod_{a=1}^{n_c} v_a \label{eq:attr_comb}
\end{equation}
\end{tabularx}
where $n_c$ is the number of attributes of category $c$. $v_a$ is the number of possible values of attribute $a$.

\begin{figure}
    \centering
    \includegraphics[width=\linewidth]{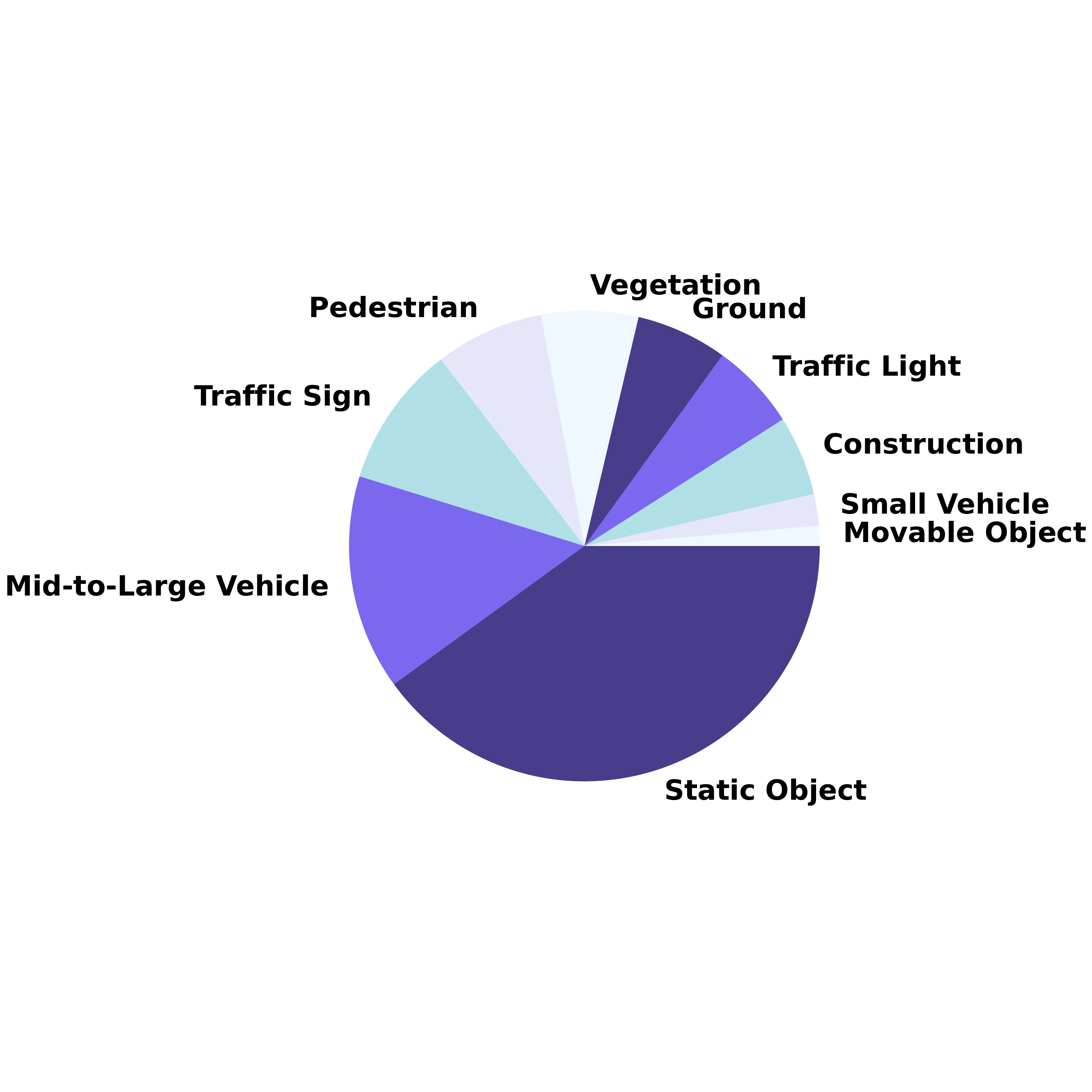}
    \caption{Pie chart showing the percentage of annotated instances for each category}
    \label{fig:n_instances_category_pie}
\end{figure}

\Cref{fig:n_instances_category_pie} shows the proportions of annotated instances of each category in \gls{car}. It can be noticed that the category with largest number of annotated instances is the ``Static Object'' category. This is largely due to their frequent existence and diversity in images of Cityscapes.

\begin{figure}
    \centering
    \includegraphics[width=\linewidth]{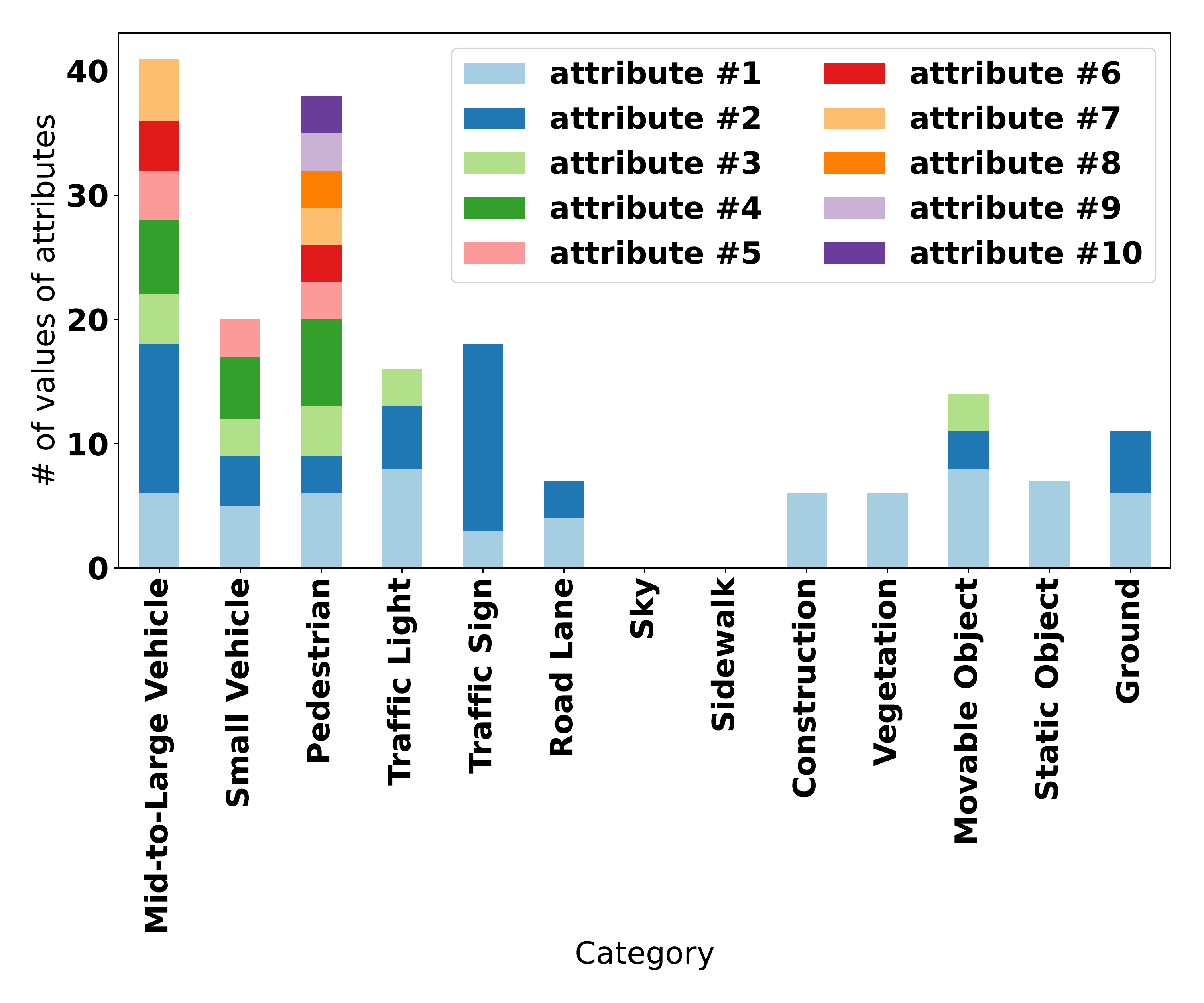}
    \caption{Bar chart showing the number of values for each attribute (distinguished by different colors) for each category (column)}
    \label{fig:n_vals_attrs_cat}
\end{figure}

\Cref{fig:n_vals_attrs_cat} depicts the number of values of each attribute in the taxonomy as some attributes may have much larger number of possible values than others. It is not possible to show the names of attributes since we have different attributes for each category. Therefore, we use numbers to distinguish between different attribute of the same category. In the Figure, each color represents one attribute of a category. Its height shows the number of possible values it may take. As it is clear from the figure, we show more interest in our labeling process to attributes related to Vehicles (Small and Large) as well as Pedestrians. These are the most important categories to understand its properties very well as it is essential for developing better self driving algorithms.

\begin{figure}
    \centering
    \includegraphics[width=\linewidth]{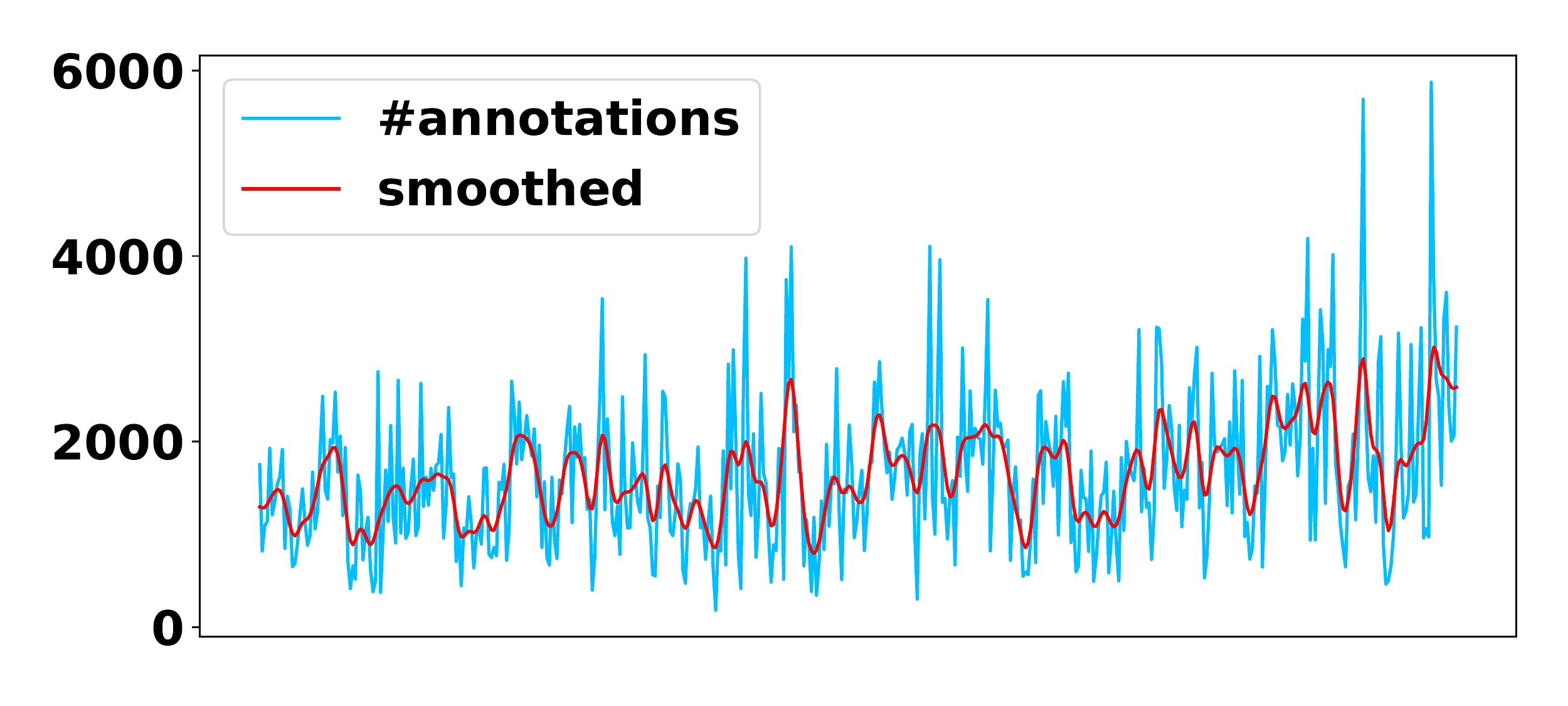}
    \caption{Number of annotations per image in CAR Dataset}
    \label{fig:n_annotations_per_image}
\end{figure}

\Cref{fig:n_annotations_per_image} discusses the number of annotations in each image. The number of objects, their categories and their possible attributes vary significantly from one image to another. Therefore, we can notice the spike and noisy output. We plot a smoothed plot as well in red. In some images, we have about 6000 attribute value. This is a huge number of annotations and it signifies the rich information the attributes contain. Adding this information to segmentation masks as well as other annotations provided by Cityscapes dataset, we end up with a rich and unique dataset that is very useful to the research community and self driving industry.

\begin{table*}
    \caption{Number of attributes per category}
    \label{tab:attributes_per_category}
    \centering
    \begin{tabularx}{\linewidth}{r||M|M|M|M|}\Xhline{4\arrayrulewidth}
    Category & \text{\#Attributes } (N_c) & \text{Avg-Vals./Attr. } (\mu) & \text{\#Combinations } (\pi) & \text{\#Annotated Instances} \\\Xhline{2\arrayrulewidth}
    Mid-to-Large Vehicle & 7 & 5.86 & 138240 & 4834 \\\hline
	Small Vehicle & 5 & 4.0 & 900 & 710 \\\hline
	Pedestrian & 10 & 3.8 & 367416 & 2445 \\\hline
	Traffic Light & 3 & 5.33 & 120 & 1959 \\\hline
	Traffic Sign & 3 & 32.67 & 3600 & 3200 \\\hline
	Road Lane & 2 & 3.5 & 12 & 0 \\\hline
	Sky & 0 & 0 & 1 & 0 \\\hline
	Sidewalk & 0 & 0 & 1 & 0 \\\hline
	Construction & 1 & 6.0 & 6 & 1798 \\\hline
	Vegetation & 1 & 6.0 & 6 & 2179 \\\hline
	Movable Object & 3 & 4.67 & 72 & 451 \\\hline
	Static Object & 1 & 7.0 & 7 & 13086 \\\hline
	Ground & 2 & 5.5 & 30 & 2051\\\Xhline{4\arrayrulewidth}
    \end{tabularx}
\end{table*}

\section{Limitations and Future Work}
\label{sec:future_work}

Even though the number of annotated instances is significant and sufficient for deep architecture to learn and predict attributes correctly, increasing the number of instances will surely help in reaching better benchmarking values. Moreover, we can focus on more types of objects or more attributes. For example, Cityscapes does not have segmentation masks for road lanes. Even though we have them in our taxonomy, we cannot label instances of those types. 

As a future work, we are planning to do the following. First, we want to increase the number of annotated instances. This will help in developing better learning algorithms that can make use of such enormous data without being susceptible to overfitting. Second, we would like to segment some categories (such as Road Lanes) that are not labeled in Cityscapes and then perform attributes labeling over them. This will help in better understanding of scenes and developing better self driving algorithms as we would have much more important features about the scene.

\section{Conclusion}
\label{sec:conclusion}

In this paper, we have discussed a novel dataset called Cityscapes Attributes Recognition (CAR). CAR complements the already existing Cityscapes dataset by adding attributes to each object in each image of the dataset. This leads to a complete understanding of a scene, which is vital in some situations of autonomous vehicles. For instance, a drive-less vehicle need to understand the expected behavior of nearby pedestrians. The dataset contains $32,729$ instances. The dataset is divided into

Having larger dataset for attributes prediction that solely focus on categories related to driving is essential for the field of autonomous vehicles. In addition, it will lead to a more sustainable and reliable algorithms for computer vision.


\section{Acknowledgement}
\label{sec:ack}
The authors would like to thank Felix Lau for his significant help in conducting this work.

{\small
\bibliographystyle{./others/ieee_fullname}
\bibliography{./main}
}

\newpage

\newpage

\appendix

\section{Attributes Taxonomy}\label{sec:supp_taxonomy}

Here, we are going to list the different possible categories (classes) then for each provide possible attributes and values it can take. We are dividing them based on whether we need to differentiate instances of each type (things) or not (stuff). For each attribute, we mention a proposed name for it, then some possible values it can take followed by a short description.

\subsection{Things}
\subsubsection{Mid-to-Large Vehicle}
This class contains all possible medium to large vehicles, which basically include all vehicles with 4 wheels. It doesn't include bicycles, motorcycles and any other small or portable vehicles.
\begin{itemize}
    \item Visibility: ranges from $0\%$ to $100\%$ with a step of $20\%$; how much of area is visible?
    \item Form: Sedan / Hatchback / SUV / Sport / Van / Pickup / Truck / Trailer / Bus / School Bus / On Rails; on rails includes Tram or Train
    \item isTowing: Yes / No; is it towing something behind it?
    \item Type: Police / Ambulance / Fire / Construction / Neither; what is this vehicle used for?
    \item lightsStatus: On / Off / Emergency; whether the lights are on or off and if there is any emergency light that is on, if the vehicle is a school bus then the emergency flag would imply that the stop sign is on.
    \item Status: Moving / Stopped / Parked; whether it is parked, stopped in the road or moving.
    \item Direction: Forward / Backward / Left / Right; whether this vehicle is forward facing, backward facing ...
\end{itemize}

\subsubsection{Small Vehicle}
It includes the remaining small or portable vehicles such as bicycles, motorcycles, scooters, ... etc.
\begin{itemize}
    \item Type: Bicycle / Motorcycle / Flat Drivable Surface / Other; Other class may include things such as scooters.
    \item Status: Moving / Stopped / Parked; whether it is moving or not. 
    \item hasRider: Yes / No; whether there is a rider or not.
    \item Direction: Forward / Backward / Left / Right; whether this vehicle is forward facing, backward facing ...
    \item hasSidecar: Yes / No; some vehicles such as motorcycles may have sidecars. This can be used for bicycles as well with an additional cart or something else.
\end{itemize}

\subsubsection{Pedestrian}
\begin{itemize}
    \item Age: Adult / Child.
    \item Type: Police Officer / Construction Worker / Neither.
    \item Activity: Sitting / Lying Down / Standing / Walking / Running / Riding.
    \item isUsingVehicle: Yes / No; whether the pedestrian is using a bicycle, motorcycle, scooter, wheelchair or something else.
    \item isPushingOrDragging: Yes / No; whether the pedestrian is pushing something in front of him / her or pulling something behind.
    \item isCarrying: Yes / No; whether the pedestrian is carrying anything (a child, a backpack). It is important to notice people with difficulties to move easily.
    \item isHearingImpaired: Yes / No; not actually due to disability, but because of wearing headphones or any item that may prevent the pedestrian from hearing the surrounding environment.
    \item isBlind: Yes / No.
    \item isDisabled: Yes / No; any kind of disability except for being blind or hair impaired.
\end{itemize}

\subsubsection{Animal}
\begin{itemize}
    \item isOnGround: Yes / No.
    \item isMoving: Yes / No.
\end{itemize}

\subsubsection{Traffic Light}
\begin{itemize}
    \item Type: Circle / Forward / Right / Left / U-Turn / Pedestrian / Unknown.
    \item Status: Green / Yellow / Red / Black; for some traffic lights such as pedestrian Green would represent pedestrians can cross and Red for no. Black is used to represent a disabled / non-working traffic light.
    \item isFlashing: Yes / No.
\end{itemize}

\subsubsection{Traffic Sign}
\begin{itemize}
    \item isElectronic: Yes / No.
    \item Type: Stop / Speed Limit / Construction / Other.
    \item Value: varies; for speed limit it contains the value.
\end{itemize}

\subsubsection{Road Lane}
\begin{itemize}
    \item Type: Solid / Broken / Other.
    \item Color: White / Yellow.
\end{itemize}

\subsection{Stuff}
\subsubsection{Sky}

\subsubsection{Sidewalk}

\subsubsection{Construction}
\begin{itemize}
    \item Building.
    \item Wall.
    \item Fence.
    \item Bridge.
    \item Tunnel.
\end{itemize}

\subsubsection{Vegetation}
\begin{itemize}
    \item Type: Tree/Hedge/Small Bush/All other kinds of vertical vegetation/other.
\end{itemize}

\subsubsection{Movable Objects}
\begin{itemize}
    \item Traffic Cones.
    \item Debris.
    \item Barriers.
    \item Pushable / Pullable.
    \item Umbrella.
    \item Other
\end{itemize}

\subsubsection{Static Objects}
\begin{itemize}
    \item Bicycle Rack.
    \item Pole.
    \item Rail Track.
\end{itemize}

\subsubsection{Ground}
\begin{itemize}
    \item Terrain. 
    \begin{itemize}
        \item Type: Grass/Soil/Sand/other; all kinds of horizontal vegetation, soil or sand.
    \end{itemize}
    \item Road.
    \item Parking lots and driveways.
\end{itemize}


\section{Discussion about best dataset to choose} \label{sec:supp_discussion_datasets}
\subsection{NuScenes}
NuScenes is the best competitive dataset for our evaluation. It has some attributes, although not many. Example of attributes are shown in \Cref{tab:attribute_nuscenes}.

\begin{table}
    \caption{Attributes for NuScenes}
    \label{tab:attribute_nuscenes}
    \centering
    \resizebox{\linewidth}{!}{%
    \begin{tabular}{r|c|c|c|c|}
    \hline
         & Visibility & Activity & Has Rider & Posture \\\hline
        Classes & All & 4wheels vehicles & \multiline{c}{Bicycles\\Motorcycles\\Portable vehicles} & Humans\\\hline
        Values & \multiline{c}{$<41\%$,\\ 41\%-60\%,\\ 61\%-80\%,\\ $>80\%$} & \multiline{c}{Parked\\Stopped\\Moving} & \multiline{c}{Yes\\No} & \multiline{c}{Sitting or Lying down\\Standing\\Moving}\\\hline
    \end{tabular}%
    }
\end{table}

\begin{itemize}
    \item They have Panoptic Segmentation for LiDAR data, but they do not have it for the images.
    
    \item They also released a new dataset NuImages (which is a standalone dataset that only has images -- without LiDAR). It has 93k images (chosen from the total 1.4M). It has different attributes taxonomy compared to NuScenes itself. Examples of attributes are shown in \Cref{tab:attributes_nuimages}.
    \begin{table}
    \caption{NuImages attributes examples}
    \label{tab:attributes_nuimages}
    \centering
    \begin{tabularx}{\linewidth}{C|Y|Y}
    \hline
        \normalsize Attribute & Freq. & \multiline{c}{Ratio} \\\hline
        cycle.with\_rider & $8,075$ & $22.4$ \\
        cycle.without\_rider & $28,036$ & $77.6$ \\
        pedestrian.moving & $102,066$ & $61.4$ \\
        pedestrian.sitting\_lying & $16,378$ & $9.9$ \\
        pedestrian.standing & $47,702$ & $28.7$ \\
        vehicle.moving & $116,855$ & $38.3$ \\
        vehicle.parked & $162,360$ & $53.2$ \\
        vehicle.stopped & $25,775$ & $8.5$ \\
        vehicle\_light.emergency.flashing & $33$ & $18.0$ \\
        vehicle\_light.emergency.not\_flashing & $150$ & $82.0$ \\
        vertical\_position.off\_ground & $33$ & $12.9$ \\
        vertical\_position.on\_ground & $222$ & $87.1$\\\hline
    \end{tabularx}%
\end{table}
    
\end{itemize}

\subsection{Cityscapes}
\begin{itemize}
    \item Much more highly cited.
    \item It spans an enormous amount of locations (50 cities). This provides a rich and diverse amount of images.
    \item Even though the number of images is lower than NuScenes, but it's significantly more diverse.
    \item If we decided to work with them, we can provide accurate segmentation masks for the 20k remaining images. We may use a pre-labeling architecture to label them (since we have a rough estimation) and then have it reviewed.
    \item In the Cityscapes website, they provide some contributions of third parties who made more labeling of the data (for example 3D Boxes). If we go with Cityscapes, we can have better visibility for our contribution through their website.
\end{itemize}

\subsection{Some week aspects of current datasets}
\begin{itemize}
    \item They do not focus on traffic signs and traffic lights. An autonomous vehicle should finally be able to deal with these tricky and difficult classes. It may be segmented in some datasets but it is not recognized as instances and no attributes are labeled for them.
\end{itemize}


\section{Annotations statistics}
\label{sec:supp_annotations_statistics}

\begin{itemize}
    \item The finely-annotated images of Cityscapes has 5k images.
    
    \item \Cref{tab:cityscapes_annotations_original} shows the original distribution of different categories in the finely-annotated images.
    \begin{table}%
        \caption{Annotations in Cityscapes}\label{tab:cityscapes_annotations_original}
        \centering%
        \begin{tabularx}{\linewidth}{Z|L}%
        \hline%
        Categories&\text{\# occurrences}\\%
        \hline\hline%
        sky&2943\\%
        \hline%
        road&3118\\%
        \hline%
        building&7141\\%
        \hline%
        sidewalk&7132\\%
        \hline%
        vegetation&15022\\%
        \hline%
        static&38430\\%
        \hline%
        bridge&392\\%
        \hline%
        pole&42904\\%
        \hline%
        traffic light&10237\\%
        \hline%
        cargroup&1895\\%
        \hline%
        person&17994\\%
        \hline%
        car&27155\\%
        \hline%
        traffic sign&20868\\%
        \hline%
        bicyclegroup&816\\%
        \hline%
        rider&1807\\%
        \hline%
        persongroup&927\\%
        \hline%
        dynamic&3491\\%
        \hline%
        fence&2467\\%
        \hline%
        terrain&4457\\%
        \hline%
        license plate&5424\\%
        \hline%
        ego vehicle&2975\\%
        \hline%
        out of roi&2975\\%
        \hline%
        bicycle&3729\\%
        \hline%
        ground&1839\\%
        \hline%
        bus&385\\%
        \hline%
        parking&1064\\%
        \hline%
        ridergroup&11\\%
        \hline%
        caravan&61\\%
        \hline%
        train&171\\%
        \hline%
        motorcycle&739\\%
        \hline%
        tunnel&29\\%
        \hline%
        polegroup&309\\%
        \hline%
        wall&1626\\%
        \hline%
        trailer&76\\%
        \hline%
        truck&489\\%
        \hline%
        rail track&112\\%
        \hline%
        guard rail&78\\%
        \hline%
        rectification border&4392\\%
        \hline%
        motorcyclegroup&9\\%
        \hline%
        truckgroup&1\\%
        \hline%
        \end{tabularx}%
        \end{table}
    
    \item Since the category definition in our case is different, we have converted them based on our category definition to be as in \Cref{tab:mapped_cityscapes_annotations}.
    
    \begin{table}%
    \caption{Cityscapes annotations after mapped to our taxonomy of categories}
    \label{tab:mapped_cityscapes_annotations}
    \centering
    \begin{tabularx}{\linewidth}{C|M|M}%
    \hline%
    \normalsize Category&\text{\# occurrences}&\text{\# attributes}\\%
    \hline\hline%
    Mid{-}to{-}Large Vehicles&33208&7\\%
    \hline%
    Small Vehicles&5293&6\\%
    \hline%
    Pedestrian&20739&9\\%
    \hline%
    Animal&0&2\\%
    \hline%
    Traffic Light&10237&3\\%
    \hline%
    Traffic Sign&20868&3\\%
    \hline%
    Road Lane&0&2\\%
    \hline%
    Sky&2943&0\\%
    \hline%
    Sidewalk&7132&0\\%
    \hline%
    Construction&16125&1\\%
    \hline%
    Vegetation&15022&0\\%
    \hline%
    Movable Objects&3491&1\\%
    \hline%
    Static Objects&81643&1\\%
    \hline%
    Ground&10590&1\\%
    \hline%
    \end{tabularx}%
    \end{table}
    
\end{itemize}

\begin{figure}
    \centering
    \includegraphics[width=\linewidth]{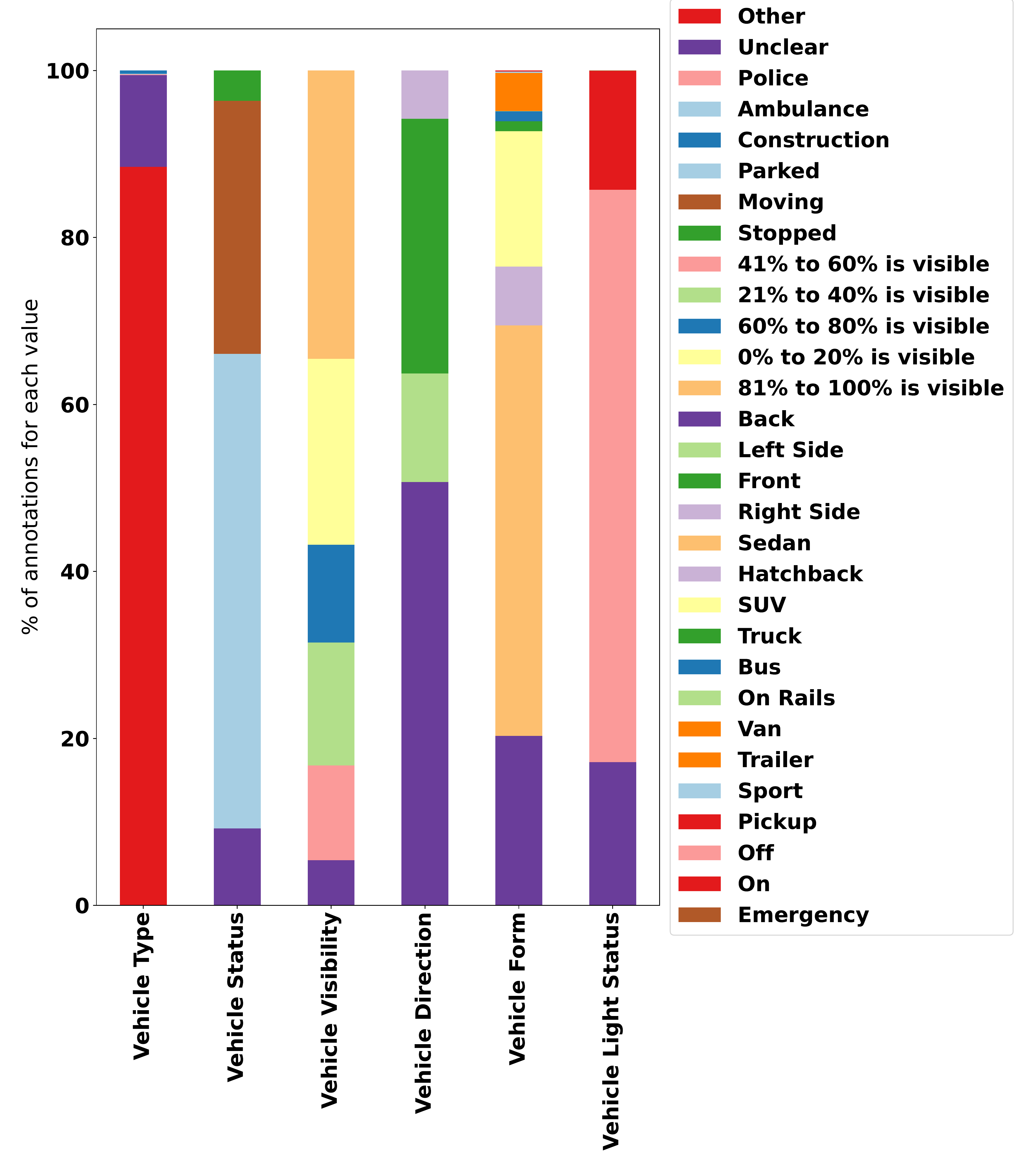}
    \caption{Large Vehicle}
    \label{fig:large_vehicle}
\end{figure}

\begin{figure}
    \centering
    \includegraphics[width=\linewidth]{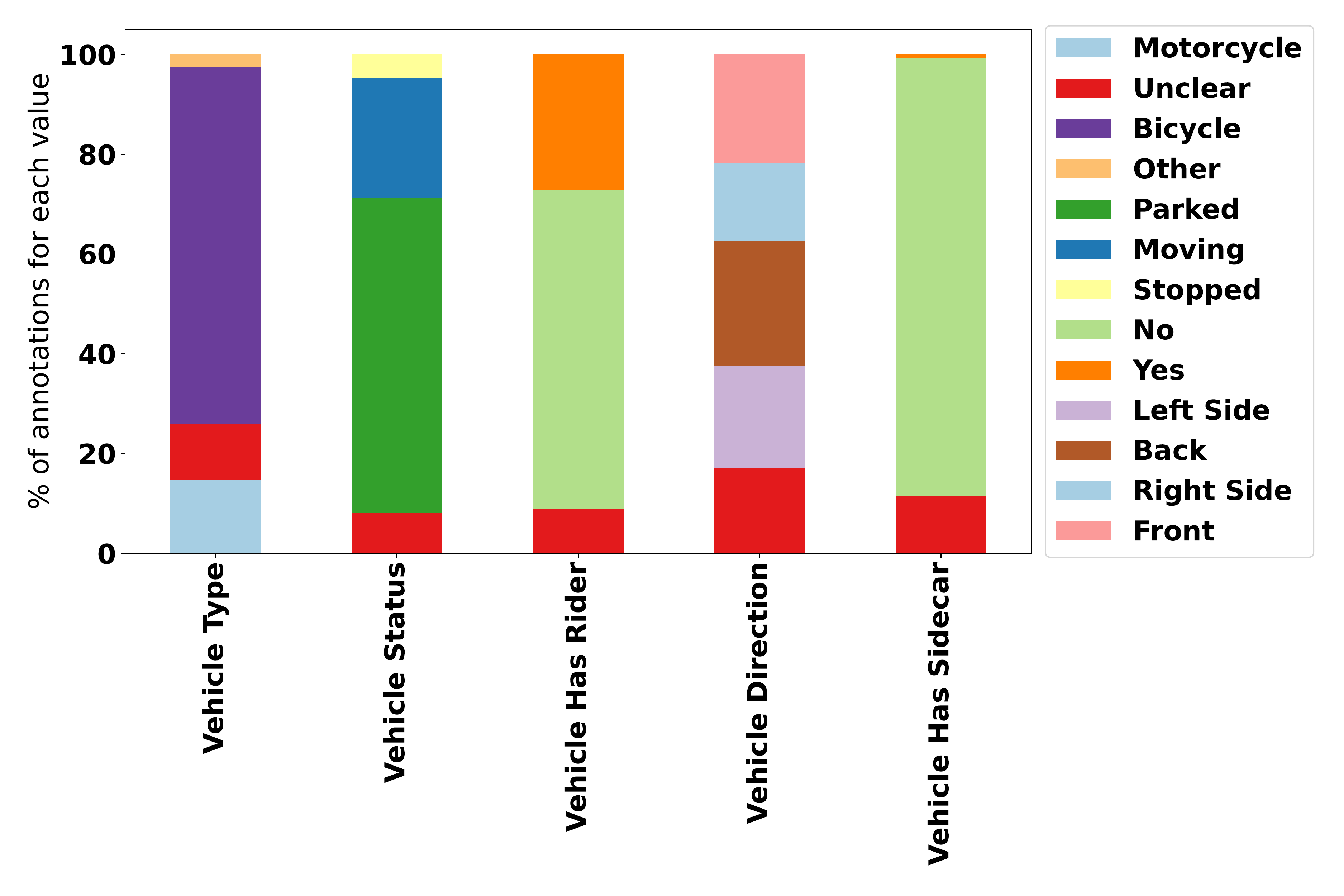}
    \caption{Small Vehicle}
    \label{fig:small_vehicle}
\end{figure}

\begin{figure}
    \centering
    \includegraphics[width=\linewidth]{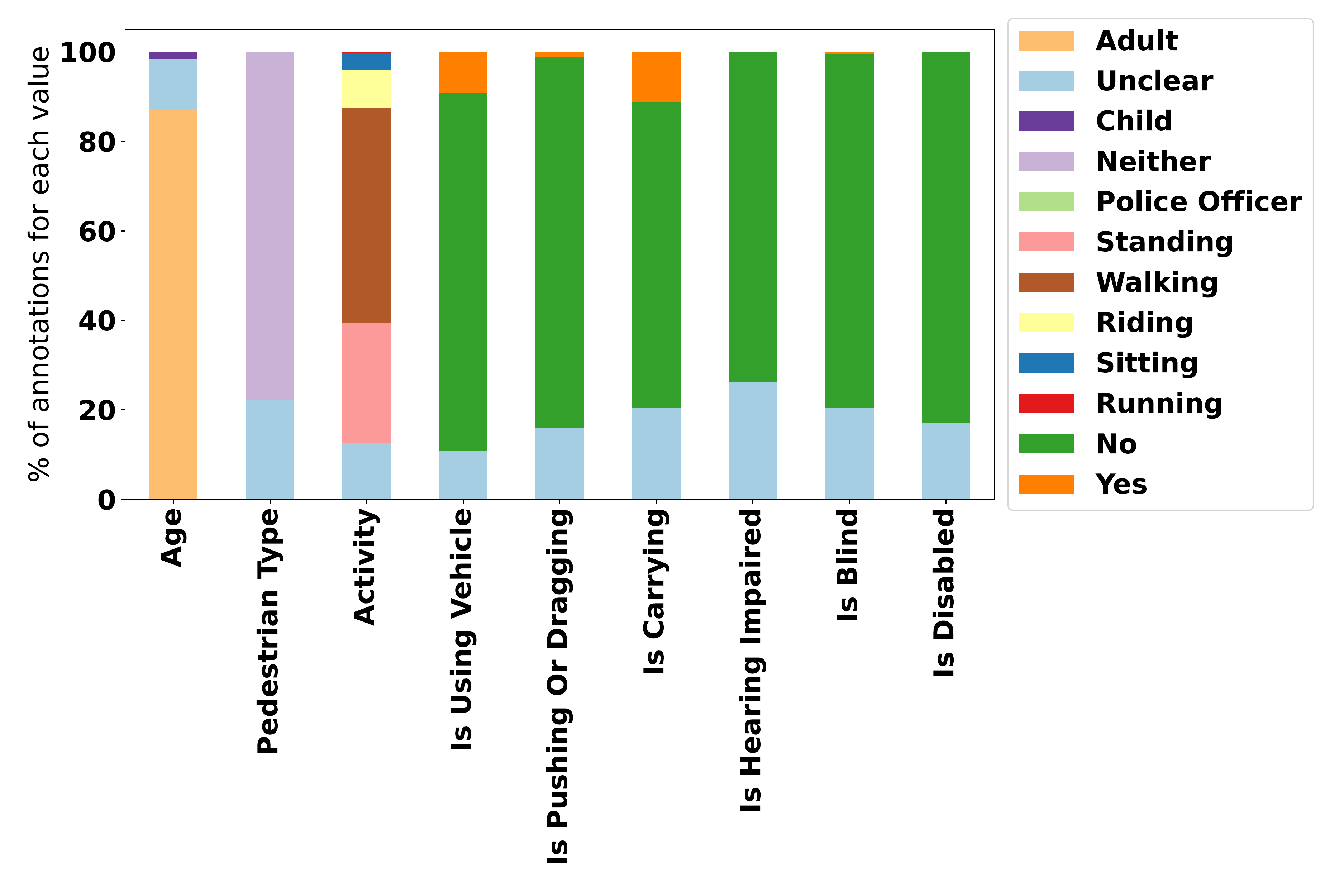}
    \caption{Pedestrian Vehicle}
    \label{fig:pedestrian_vehicle}
\end{figure}

\end{document}